\documentclass[conference]{IEEEtran}
\IEEEoverridecommandlockouts
\renewcommand{\baselinestretch}{0.901}
\usepackage{amsbsy,amsfonts,amssymb,amsmath,color}
\usepackage{graphics}
\usepackage{graphicx}
\usepackage{epstopdf}
\usepackage{textcomp}
\usepackage{multirow}
\usepackage[export]{adjustbox}
\usepackage{bm}
\usepackage{lipsum}
\usepackage{url}
\usepackage{cite}

\def\b0{{\bf 0}}

\newcommand\blfootnote[1]{%
	\begingroup
	\renewcommand\thefootnote{}\footnote{#1}%
	\addtocounter{footnote}{-1}%
	\endgroup
}

\makeatletter
\def\footnoterule{\relax%
    \kern-5pt
    \hbox to \columnwidth{\hfill\vrule width 1\columnwidth height 0.4pt\hfill}
    \kern4.6pt}
\makeatother

\begin{document}

\author{\IEEEauthorblockN{\large Anil Yesilkaya, Onur Karatalay, Arif Selcuk Ogrenci, Erdal Panayirci }\\
	
	\IEEEauthorblockA{ Kadir Has University \\ Istanbul, Turkey\\
		 \{\ anil.yesilkaya, onur.karatalay, ogrenci, eepanay \}@khas.edu.tr\\
		}}

\title{Channel Estimation for Visible Light Communications Using Neural Networks}
\maketitle

\begin{abstract}
Visible light communications (VLC) is an emerging field in technology and research. Estimating the channel taps is a major requirement for designing reliable communication systems. Due to the nonlinear characteristics of the VLC channel those parameters cannot be derived easily. They can be calculated by means of software simulation. In this work, a novel methodology is proposed for the prediction of channel parameters using neural networks. Measurements conducted in a controlled experimental setup are used to train neural networks for channel tap prediction. Our experiment results indicate that neural networks can be effectively trained to predict channel taps under different environmental conditions.
\blfootnote{This work  is supported by COST-TUBITAK Research Grant
	No: 113E307. \\ }
\end{abstract}

\vspace{1mm}



\section{Introduction}
Optical wireless communications (OWC) has attracted great attention of researchers and engineers recently. The spectrum bottleneck associated with great demand of high data rates for mobile data usage pushes researchers to develop new technologies for wireless communications such as, millimeter wave, free space optical, underwater acoustic communications. Visible light communications (VLC) is one of the promising technology that is considered for 5G or further communications standards. VLC has many advantages over radio frequency systems (RF) approximately 10.000 times bigger and unregulated bandwidth, low cost for deployment, higher security and lower interference from other RF devices \cite{haas}.

OWC comprises VL (visible light) and IR (infra-red) regions of the spectrum as indoor/outdoor wireless communications medium. Visible light communications (VLC) is a branch of OWC operating in the VL (390nm-750nm) band. Intensity Modulation / Direct Detection (IM/DD) method is accepted as the most applicable modulation technique to transmit data over visible light. In IM/DD data are coded on the small intensity fluctuations. At the receiver, photo-detectors capture fluctuations and convert them to digital data \cite{4785281}. A proper channel model is one of the most important components to have robust, error-free and reliable wireless communications systems. Despite the ever increasing popularity of the visible light communications, there is a lack of a proper VLC channel model. Obtaining an analytical expression for the channel is almost impossible due to the unpredictable changes in the environment. At this point, we propose that artificial neural networks (ANN) can provide a practical and reliable approach. Artificial neural networks are quite powerful tools to model the relationship between inputs and outputs of the system and they are quite useful when that relationship is non-linear. In this paper, we used multi-layer perceptron (MLP) network to construct a real time VLC channel estimator to obtain channel taps under different environmental conditions with high accuracy. We constructed a realistic indoor environment in the laboratory and used real materials to investigate effects of the surface types having different reflectance values. In the sequel, channel taps are estimated by using asymmetrically clipped optical orthogonal frequency division multiplexing (ACO-OFDM) for real life scenarios. The learning phase based on ANN has given us a model to estimate the channel taps. Then different groups of materials having different reflectance values have been used to test the validity of that model. Measurement results indicate that the model is capable of calculating the channel taps with an average accuracy higher than 97.7\% in the training. The major contribution of the paper can be stated as follows: the real time channel model in VLC can be constructed using artificial neural networks based on a set of minimal measurements. Results show that, even with limited number of parameters and experiments, convenient channel models for VLC can be obtained.

The rest of the paper is organized as follows: In Section II, we describe the visible light communications channel properties and its challenges. In Section III, we describe the methodology adopted for VLC channel estimation. In Section IV, we present MLP channel estimator and its performance. Finally, we conclude the paper in Section V.

\section{Challenges in VLC Channel Modeling}
 Reflection and refraction patterns are already well defined for daily life materials however, dynamic parameters are affecting the VLC channel (e.g. moving objects and people, fluctuations in noise sources, unknown reflections of mixed type materials etc.) which complicate the derivation of an analytical expression for the channel model. Obtaining proper channel model ensures designing reliable and robust communication systems. Yet, in the literature most of the researches are using infra-red (IR) channel models or simple additive white Gaussian noise (AWGN) channel to model VLC environment \cite{6415964,6821328}. In \cite{5682214}, IR sources are defined as monochromatic where white LED's are considered as wide-band sources (380nm-780nm) intrinsically. It could be seen that wavelength dependent VL channel models are required. Previous studies about frequency selective multi-path VLC channel modeling are based on numerically computed non-sequential ray-tracing approach. For higher data rates VLC channel has frequency selective behavior \cite{7339420}. Frequency selectivity basically means that channel acts as a simple FIR filter described by coefficients which are called "channel taps" in the communication literature. Obtaining channel taps brings great control over distortion cancellation in the received signal. These channel taps are used to model channel impulse response (CIR) which can be expressed as attenuations and time delays as,
 \begin{equation}
 h(t) = \sum_{i=1}^{N}{P_{i}\delta(t - \tau_{i})}
 \end{equation}
 where $ P_{i} $ is the power and  $ \tau_i $ is the  propagation time of the {\it $i^{th}$} ray, $ \delta $ is the Dirac delta
 function and $ N $ is the number of rays received in the detector.
 Based on the obtained CIR, we can further define the fundamental
 channel characteristics. 
 
 Channel DC gain $ (H_0) $ is one of the
 most important features of the VLC channel. It determines the
 achievable signal-to-noise ratio (SNR) for fixed transmitter power.
 The delay profile is composed of dominant multiple line of sight (LOS) links and
 less number of non-line of sight (NLOS) delay taps. The temporal dispersion of a power
 delay profile can be expressed by the mean excess delay ($ \tau_{0}
 $) and the channel root-mean-square (RMS) delay spread ($ \tau_{\text{RMS}}
 $). These parameters are given by \cite{7339420},

\begin{equation}
\int_0^{T_r}{h(t)dt} = 0.97 \int_0^{\infty}{h(t)dt}
\end{equation}

\begin{equation}
\tau_0 = \frac{\int_0^{\infty}t \times h(t)dt
}{\int_0^{\infty}h(t)dt }
\end{equation}

\begin{equation}
\tau_{\text{RMS}} = \sqrt{\frac{\int_0^{\infty}(t-\tau_0)^2 h(t)dt }{\int_0^{\infty}h(t)dt}}
\end{equation}

\begin{equation}
H_0 = \int_{-\infty}^{\infty}h(t)dt
\end{equation}

From (2) it can be seen that 97 percent of the power of the CIR is contained in the [0,$T_r$] interval. In our experiments, 512kHz bandwidth is selected for data transmission and up to 2 channel taps are enough to model the channel adequately \cite{5206382}.

\section{Methodology}
\begin{figure}[ht]
	\centering
	\includegraphics[width=230px, height=45px]{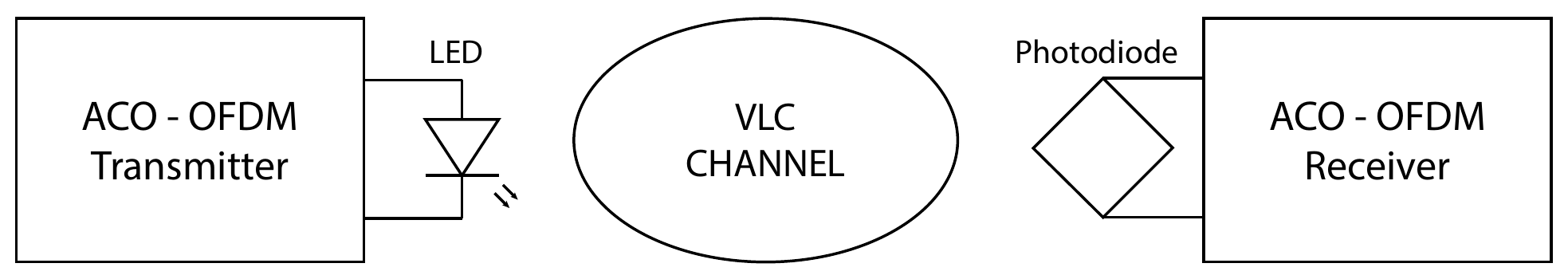}
	\caption{ACO OFDM block diagram for VLC channel estimation}
	\label{f1}
\end{figure}

The block diagram of the transmitter and receiver part for ACO-OFDM based IM/DD (intensity modulation/direct detection) system is shown in Fig. \ref{f1}. In IM/DD method, commercial LED's are used as a transmitter by carrying information in the intensity of light where photo diodes are used as a receiver to detect small fluctuations in the light intensity.

In the transmitter part, user generated bit stream is modulated and carried by LEDs where VLC channel part conveys various disturbances such as ambient lights, reflections, refractions and obstructions. Lastly, at the receiver, channel coefficients are estimated by using already known signals (pilot symbols). Then, estimated channel coefficients are used in the MLP training to predict the channel taps in different environments without using further knowledge.

\subsection{Selection of Materials}
The channel model (taps) heavily depend on the surface types of the environment because receiver captures reflected rays. For this reason the neural neural network model should include surface materials covering a wide range of reflectivity. Materials in the experiments are selected from NASA's spectral database \cite{aster} and realistic indoor configuration is created in the laboratory setup. Materials used in the experiments and their relative reflectances are shown in Fig. 2 and in Table I respectively.

\begin{figure}[h]
	\centering
	\includegraphics[width=\columnwidth, height=90.6px]{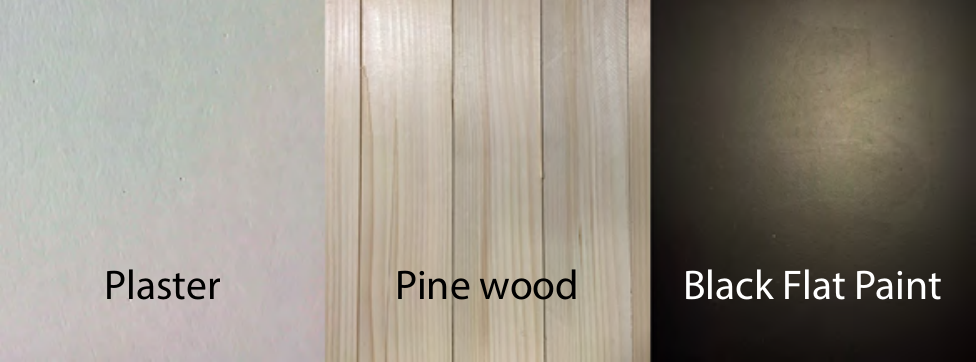}
	\caption{Photographs of materials used in the experiments.}
	\label{f6}
\end{figure}

\begin{figure}[h]
	\centering
	\includegraphics[width=80mm, height=45mm]{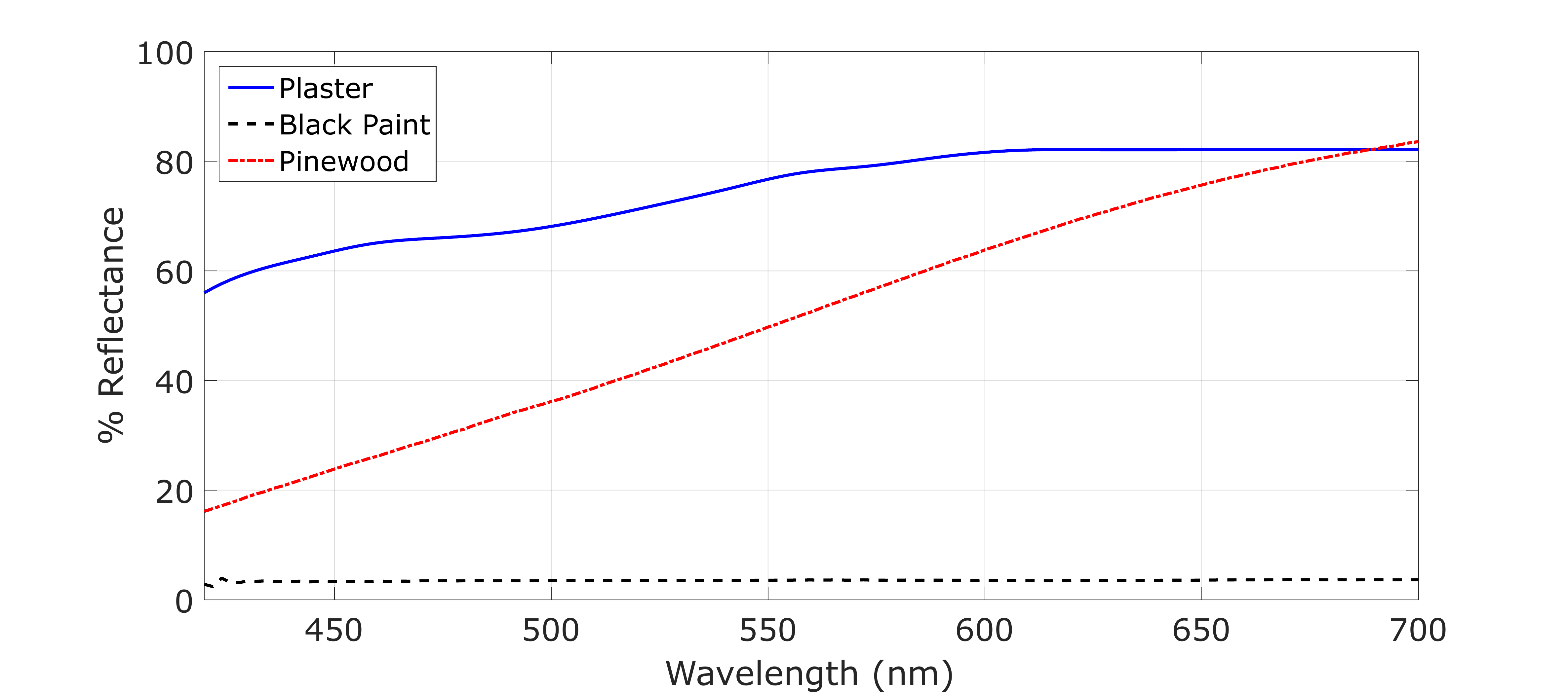}
	\caption{Relative reflectance values for the materials}
	\label{f66}
\end{figure}

From Fig. 3, it can be deduced that the plaster has the highest average reflectance where the pine wood and black flat paint have intermediate and lowest average reflectance values respectively, over the VL band. Average reflectivity over 420nm-700nm band can be calculated from \% Reflectance $\bm{\rho}$ and normalized spectral power distribution vectors $\bm{P_{N}}$ as,
\begin{equation}
	\rho_{\text{avg}}=\bm{\rho}\times\bm{P_{N}}=\frac{1}{P_{\text{Total}}}\sum_{k=1}^{M}\rho_kP_k
\end{equation}
where $\rho_{\text{avg}}$ represents the average reflectance values for VL band and total spectral power distribution $P_{\text{Total}}$ can be calculated from the areas under the curves in Fig. 4 as $\sum_{l=1}^{M}P_l $ where $M$ and $P_l$ are the number of sample points between 420nm and 700nm and discrete spectral power distribution of the LEDs respectively.

\begin{figure}[h]
	\centering
	\includegraphics[width=\columnwidth, height=45mm]{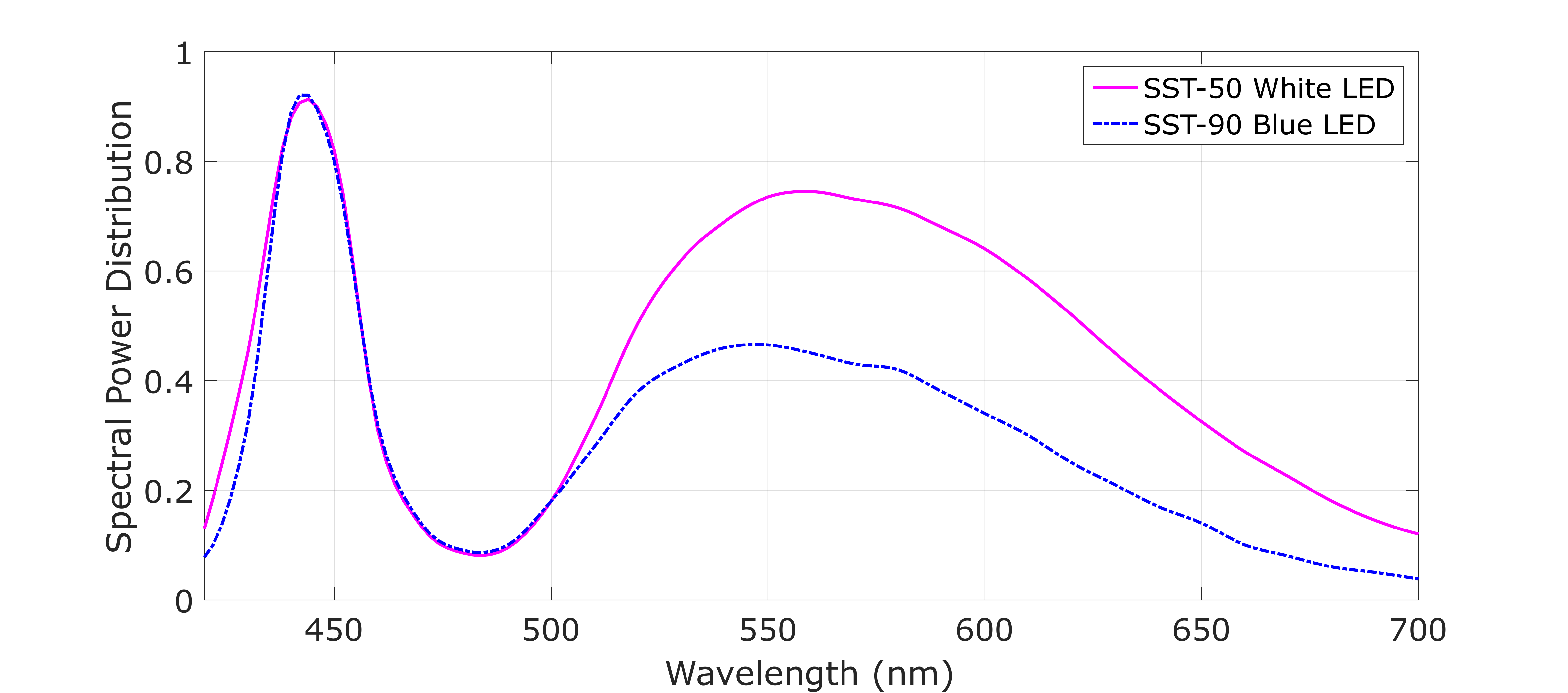}
	\caption{Spectral power distributions of the SST-50 White and SST-90 Blue LEDs}
	\label{f77}
\end{figure}

\subsection{Selection of Light Sources and Detectors}
\begin{table*}[ht]
	\centering
	\caption{Input features used in MLP}
	\label{my-label}
	\resizebox{18cm}{!}{
		\renewcommand{\arraystretch}{1.3} 
		\begin{tabular}{|c|c|c|c|}
			\hline
			\textbf{Attributes} & \textbf{Descriptions} & \multicolumn{2}{c|}{\textbf{Values}} \\ \hline
			Reflectivity & \begin{tabular}[c]{@{}c@{}}Low: Black Flat Paint\\ Medium: Pine wood\\ High: Plaster\end{tabular} & \begin{tabular}[c]{@{}c@{}}Average Reflectance for White LED\\ Black Flat Paint: 0.0352\\ Pine wood: 0.5059\\ Plaster: 0.7489\end{tabular} & \begin{tabular}[c]{@{}c@{}}Average Reflectance for Blue LED\\ Black Flat Paint: 0.0350\\ Pine wood: 0.4541\\ Plaster: 0.7285\end{tabular} \\ \hline
			Transmitter type & \begin{tabular}[c]{@{}c@{}}White LED\\ Blue LED\end{tabular} & Average Spectral Power for White LED: 63.02 W & Average Spectral Power for Blue LED: 41.56 W \\ \hline
			LOS/NLOS & \begin{tabular}[c]{@{}c@{}}NLOS\\ NLOS+LOS\end{tabular} & \multicolumn{2}{c|}{\begin{tabular}[c]{@{}c@{}}NLOS: 45\textdegree between Rx and surface normal while Tx and Rx are directed to each other.\\ NLOS+LOS: 0\textdegree between Rx and surface normal while Tx and Rx are directed to each other.\end{tabular}} \\ \hline
			Noise Level & \begin{tabular}[c]{@{}c@{}}External light sources are OFF\\ Only one external light source is ON\\ External light sources are ON\end{tabular} & \multicolumn{2}{c|}{\begin{tabular}[c]{@{}c@{}}External light sources are OFF: 1\\ Only one external light source is ON: 2\\ External light sources are ON: 3\end{tabular}} \\ \hline
			Distance between Tx and Rx & 20 cm to 200 cm & \multicolumn{2}{c|}{\begin{tabular}[c]{@{}c@{}}For training  data, from 20 cm to 200 cm by increments of 20 cm\\ For test data, from 20 cm to 200 cm by increments of 5 cm\end{tabular}} \\ \hline
			Receiver Gain & 10 dB to 30 dB & \multicolumn{2}{c|}{\begin{tabular}[c]{@{}c@{}}$20~\text{cm}\leq \text{distance} <40~\text{cm}$ : 10 dB\\ $40~\text{cm}\leq \text{distance} <80~\text{cm}$ : 20 dB\\ $80~\text{cm}\leq \text{distance} <200~\text{cm}$ : 30 dB\end{tabular}} \\ \hline
		\end{tabular}
	}
\end{table*}

In the experiment, two different types of light sources are used to investigate the effects of wavelengths on reflectivity. For that purpose, single chip, white and blue power LEDs are used. Spectral power distributions of these LEDs are shown in Fig. 4.

Since these power LEDs are manufactured for illumination purposes only, the bandwidth and linearity of the LEDs must be taken into consideration. Those problems will be detailed in the following part.

\subsection{Measurement Setup and Channel Estimation}
System hardware design and the ACO - OFDM transmitter structure are shown in Fig. 5.
 
 \begin{figure}[ht]
 	\centering
 	\includegraphics[width=230px, height=40px]{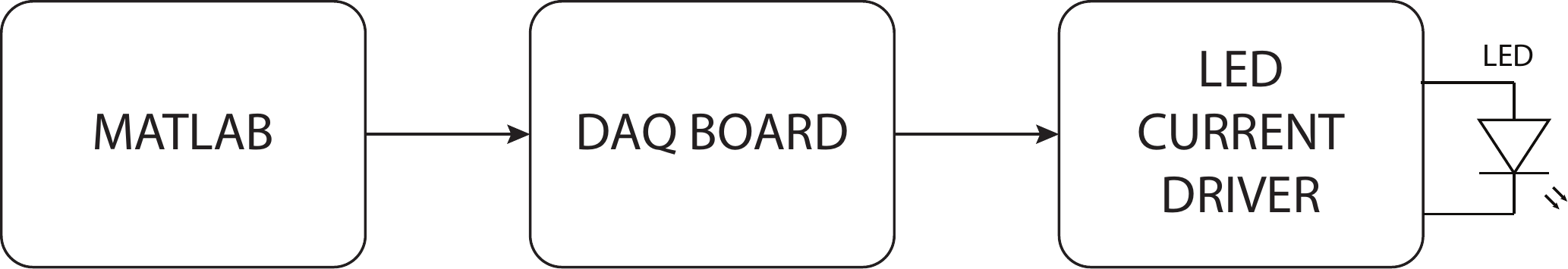}
 	\caption{Block diagram of the transmitter part}
 	\label{f2}
 \end{figure}

Data processes on the OFDM frame are executed in the computer simulation environment. State of the art digital acquisition board (DAQ) connects digital and real world together by synthesizing the digital samples using sample and hold circuit of 16 bits resolution. DAQ board has 1V/V transfer function which means that discrete samples of 1V magnitude in software can be synthesized as 1V in the real world. Generated analog signals pass from the laser diode driver with 300mA/V transfer function. For typical LEDs, illumination and current has a linear relationship within the LED's linear region. To make sure that the LEDs are operating in the linear region, the maximum transmitted signal power is limited to 1.25 Watts \cite{5962752}. 

\begin{figure}[ht]
	\centering
	\includegraphics[width=155px, height=40px]{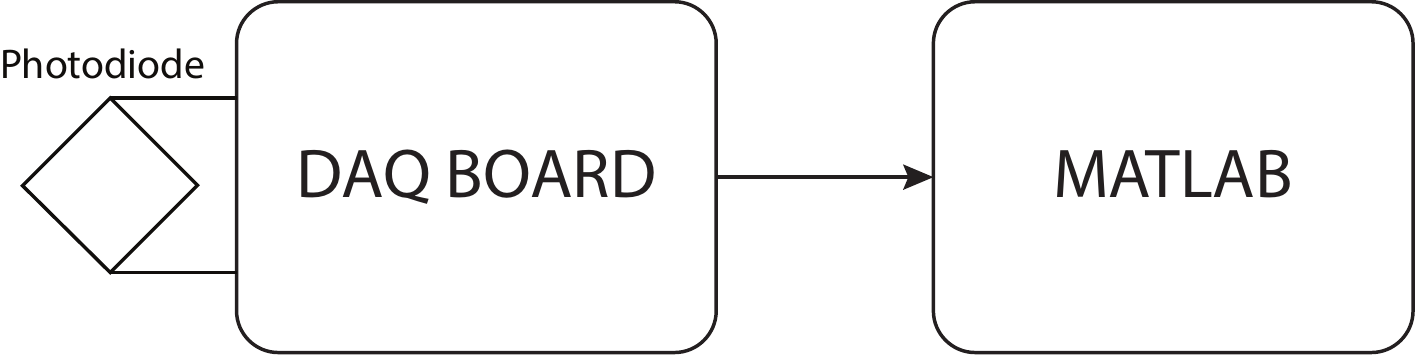}
	\caption{Block diagram of the receiver part}
	\label{f3}
\end{figure}

In the receiver part shown in Fig. 6, Silicon Based Switchable Gain Detector is used to capture very tiny alterations in the light intensity. Synchronization between the transmitter and receiver is ensured by DAQ board's built-in trigger mechanism. According to the system setup defined above, OFDM frame structure is obtained as follows. Random generated source bits are transmitted in the blocks of $T_{sym}$ duration and modulated in M-QAM modulator where they are processed parallel in further blocks with symbol duration of $T_s=T_{sym}/N$. The total number of actively used sub-carriers are represented as $N$ and for simplicity N is taken to be equal to the IFFT block size. Input signal in the frequency domain $
\textbf X=[X_0, X_1, X_2,\cdots, X_{N-1}]^T$ meets the Hermitian symmetry and only the odd indexed sub-carriers contain data where the $0^{\text{th}}$ (DC) and $(N/2)^{\text{th}}$ sub-carriers are set to zero to avoid any complex term and fulfill the Hermitian symmetry \cite{4785281,6415964}.

\begin{equation}
X[k] = \left\{\begin{array}{lll}
0 & , & \mbox{$k$ is even}\\
X^\ast_{N-k} & , & \mbox{$k$ is odd}\\
\end{array}\right.
\end{equation}
where $\ast$ denotes the complex conjugate. Lowercase letters are used for time-domain signal representations and
uppercase letters are for discrete frequency-domain signals. The resulting real, bipolar and anti-symmetric time-domain IFFT signal is given by, 
$x=[x_0,x_1,\cdots, x_{N-1}]^T$ .

\begin{equation}
x[n] = \frac{1}{\sqrt{N}} \sum_{k=0}^{N-1}X[k] e^{j\frac{2\pi kn}{N}}
\end{equation}
where $N$ is the number of points in IFFT and $X[k]$ is the $k^{\text{th}}$
sub-carrier of signal $\textbf X$ which contains already known pilot symbols for channel estimation procedure. Due to Hermitian symmetry and zero
insertion process, the number of  data symbols carried by
sub-carriers in ACO-OFDM is only $N/4$. A cyclic prefix (CP) is
then  added to the discrete time samples, where $N_{\text{CP}}$ is denoted by the length of the CP. In our experiment, $N_{\text{CP}}$ is taken as greater or equal to $ L_h$ where $L_h$ is the length of the impulse response of the optical channel.
Negative part of the signal is clipped to generate real and unipolar
signal is given by,

\begin{equation}
{ \lfloor x[n] \rfloor }_c  =
\left\{
\begin{array}{ll}
x[n]  & \mbox{if } x[n] \geq 0 \\
0 & \mbox{if } x[n] < 0.
\end{array}
\right.
\end{equation}
The clipping noise will fall only on the
even sub-carriers and will not affect the transmitted symbols carried
by odd sub-carriers. There is no need to add a DC bias to the
clipped signal in the conventional ACO-OFDM system.

\begin{table}[!t]
	\centering
	\caption{Parameters of Experiment}
	\label{table1}
	\resizebox{8cm}{!}{
		\renewcommand{\arraystretch}{1.4} 
		\begin{tabular}{|c|c|c|}
			\hline
			\textbf{Parameter} & \textbf{Description} & \textbf{Value} \\ \hline
			$N$ & OFDM frame length & 512 \\ \hline
			$K$ & Active sub-carrier number & 504 \\ \hline
			$P_s$ & Pilot Separation & 2 \\ \hline
			$L_{\text{cp}}$ & Cyclic Prefix length & 4 \\ \hline
			$C_{\text{type}}$ & Constellation type & 4-QAM \\ \hline
			$V_{\text{trigger}}$ & Triggering voltage at the receiver & 0.075V \\ \hline
			$V_{\text{bias}}$ & Bias level at the transmitter & 150mA \\ \hline
			$H_{\text{TX}}$ & Transmitter height from material & 19cm \\ \hline
			$H_{\text{RX}}$ & Receiver height from material & 19cm \\ \hline
		\end{tabular}
	}
\end{table}

At the receiver, photo-diode detects and converts
the optical signals into electrical signals. Received signal contains amplified/attenuated components as well as inter-symbol interference (ISI) and additive white Gaussian noise (AWGN). Received time-domain signal has the form of,

\begin{equation}
	y(t) = x(t) \circledast h(t) + w(t)
\end{equation}
where $\circledast$ denotes the circular convolution operation, $h(t)=[h(0),
h(1), \ldots, h(L_h-1)]^T$ is the $L$-path impulse response of the
optical channel and $w(t)$ is an AWGN that represents the
noise in the environment. Ambient noise is in the form of DC and AWGN is added in the electrical domain and
overall noise power is denoted by $\sigma_n$. 

\begin{figure}[h]
	\centering
	\includegraphics[width=160px]{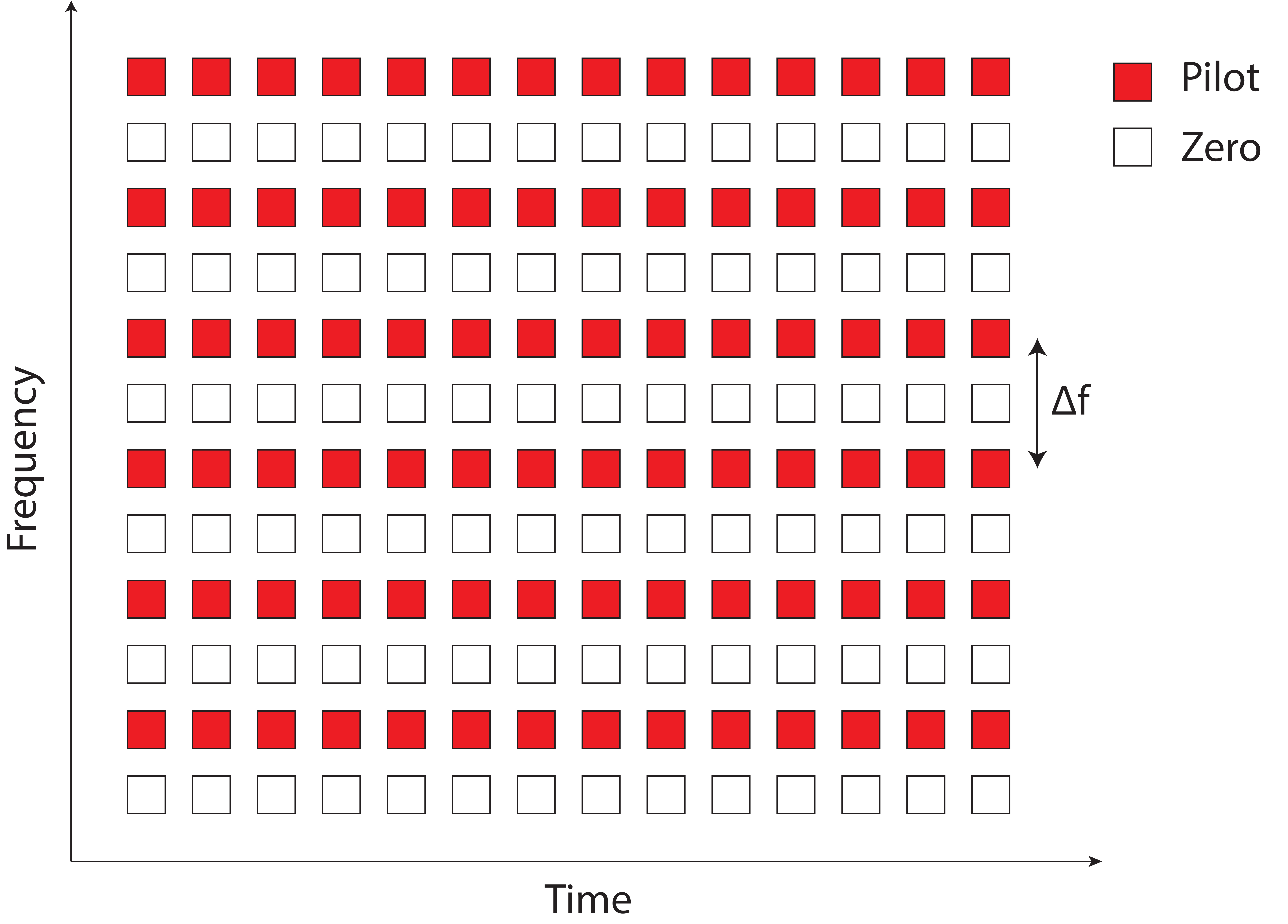}
	\caption{Comb type pilot symbol arrangement for ACO-OFDM}
	\label{f4}
\end{figure}

After detection, A/D device converts the analog signal
into the digital domain. After removing the CP, the Fast Fourier
Transform (FFT) of the received signal $y $ is taken in the computation software. Channel is estimated in frequency domain by single tap zero forcing equalizer (ZFE)  as,
\begin{equation}
	H_{\text{est}}=\mathbf{X^{-1}}\mathbf{Y}=H+\frac{W}{X}
\end{equation}
where $H_{\text{est}}$ shows $N\times1$ estimated channel taps vector in the frequency domain, $\mathbf{X}$ is the $N\times N$ diagonal pilot symbols matrix and $W$ is $N\times1$ AWGN vector in the frequency domain. Due to fact that only odd sub-carriers are modulated and even sub-carriers left as zero this is comb-type channel estimation for ACO-OFDM as shown in Fig. 7. Thus, cubic interpolation is applied to $H_{\text{est}}$ before taking IFFT and the time domain representation of the channel taps are obtained \cite{806820}. The parameters of the communication system used in the experimental setup are given in Table II.

\subsection{Multi-Layer Perceptron Structure}
Designed structure of the MLP is shown in Fig. 8 which contains one hidden layer \cite{1165576}.

\begin{figure}[h]
	\centering
	\includegraphics[width=200px, height=120px]{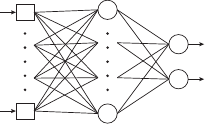}
	\caption{MLP Structure}
	\label{f5}
\end{figure}
As explained in Section II, only two channel taps are enough to model the effects in optical wireless channel. Thus, the MLP structure will have two outputs. Although there are many parameters that affect the channel directly, in this experiment six major attributes are chosen as MLP inputs which are given in Table I in detail.

The training performance of the MLP is given in Table III in terms of MSE (mean squared error) in the column "training". Another measure of performance is displayed in Table IV in the "training" column. The channel taps can be estimated with a mean percentage absolute error of less than 2.3\%. The behavior of MSE in training epochs is displayed in Fig. 9.

\begin{table}[!t]
	\centering
	\caption{MSE Performance in Training Phase}
	\label{my-label}
	\resizebox{\columnwidth}{!}{
		\renewcommand{\arraystretch}{1.5} 
		\begin{tabular}{|c|c|c|c|}
			\hline
			\textbf{Channel Tap /  Phases} & \textbf{Training} & \textbf{\begin{tabular}[c]{@{}c@{}}Test\\ (Black Flat Paint - Pine wood)\end{tabular}} & \textbf{\begin{tabular}[c]{@{}c@{}}Test\\ (Plaster - Pine wood)\end{tabular}} \\ \hline
			\textbf{h1} & $1.8\times10^{-06}$ & $3.2\times10^{-04}$  & 8.0$\times10^{-05}$ \\ \hline
			\textbf{h2} & $6.0\times10^{-07}$ & $1.5 \times10^{05}$ & $3.0 \times10^{-06}$ \\ \hline
		\end{tabular}
	}
\end{table}

For testing the trained MLP system, we have expanded the general model which depends on the following parameters: reflectivity, transmitter type, NLOS/LOS and noise levels. Two different cases are created for reflectivity values where the surface is covered with half black flat paint and half pine-wood as the first case and for the second case, the surface is covered with half plaster and half pine-wood. According to white and blue LEDs, new hybrid reflectance values are given in Table V. Parameters related to the MLP structure and training are given in the next section.

\section{Measurements and Results}
\begin{figure}[!t]
	\centering
	\includegraphics[width=90mm]{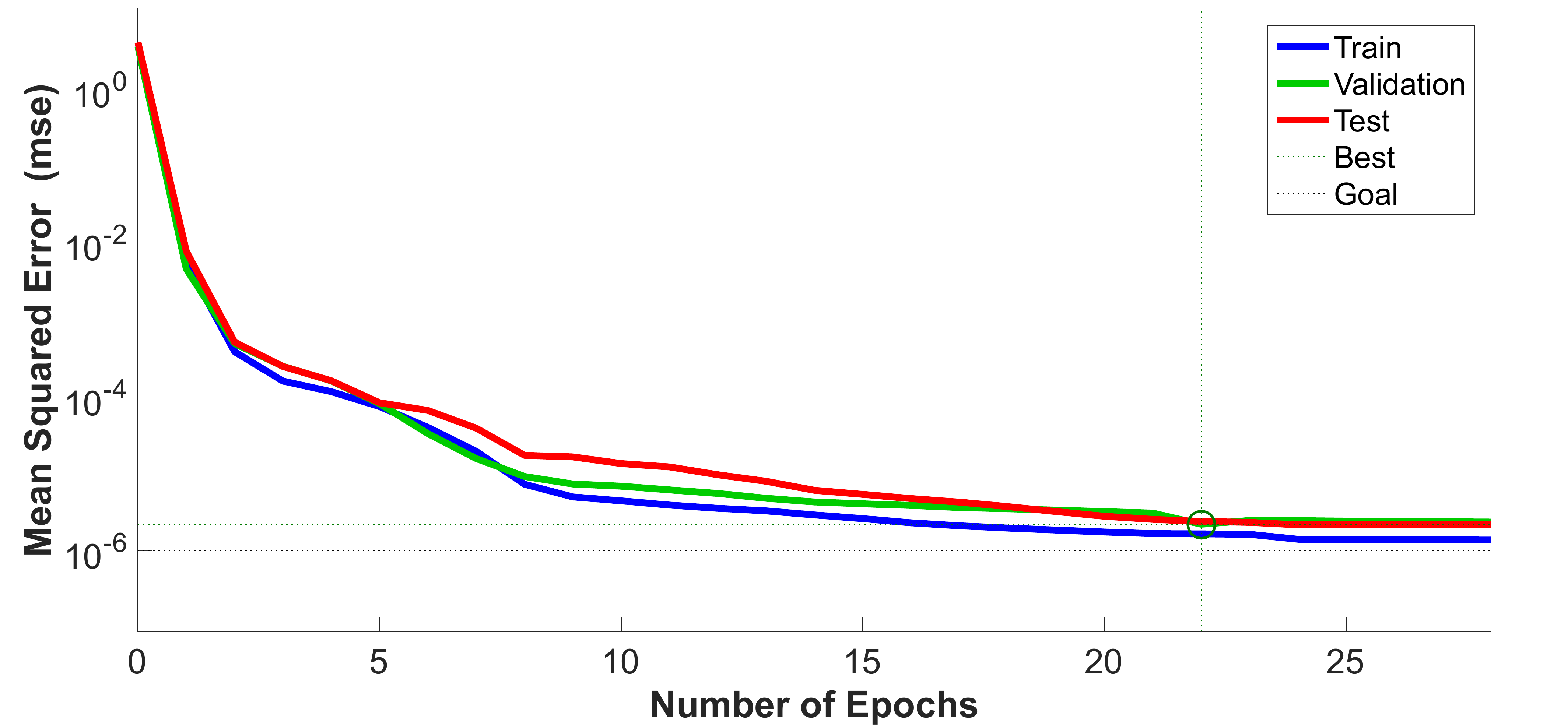}
	\caption{MSE performance of training phase}
	\label{f8}
\end{figure}
\begin{figure}[!t]
	\centering
	\includegraphics[width=85mm]{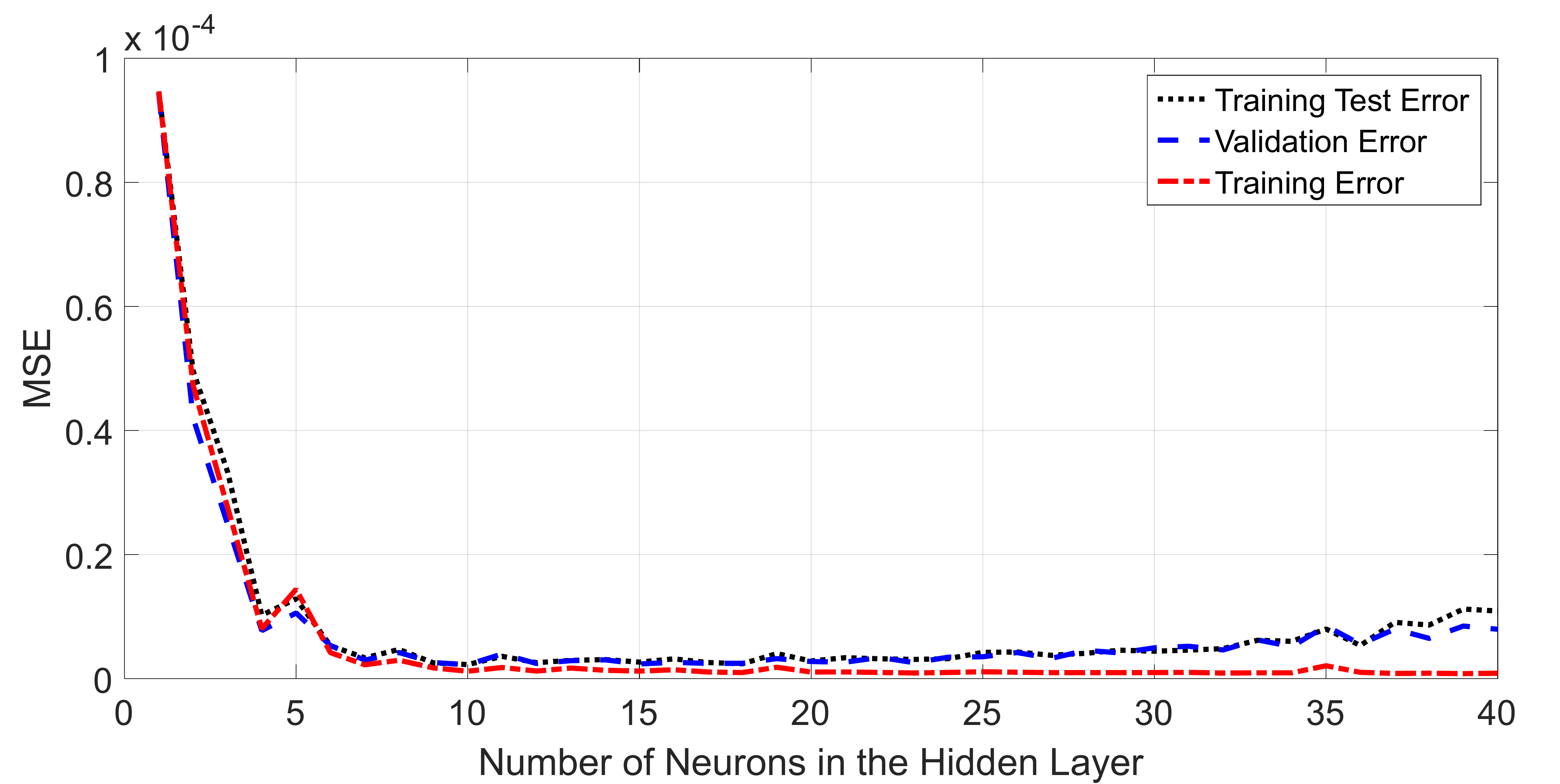}
	\caption{System performance for different number of hidden neurons}
	\label{f7777}
\end{figure}
Measurements are performed for 10 distance values in the interval of $[20~\text{cm}:200~\text{cm}]$ by increments of 20cm between the receiver and the transmitter units. Three different materials are used: plaster, pine-wood and black flat paint for high, medium, low reflectance values respectively. In the transmitter, white and blue color LEDs are used. NLOS and LOS effects are realized by 0 and 45 degrees of angle between material surface and photo-diode normal. Three different disturbance levels are created by external light sources in the environment. According to the distances between the transmitter and the receiver, power control procedure is applied at the receiver to prevent saturation. Thus, receiver gain is chosen for $20~\text{cm}\leq distance <40~\text{cm}$ interval as 10 dB, for $40cm\leq \text{distance} <80~\text{cm}$ interval as 20 dB and for $80~\text{cm}\leq \text{distance} <200~\text{cm}$ interval as 30 dB. In order to increase validity of the measurements, mean of the 10 consecutive measurements are used as estimated channel taps. In total, $3\times2\times2\times3\times10=360$ measurements are used to estimate 2 channel taps according to the method given in the Section II-C. These measurements are used for training of the MLP where tangent sigmoid function is used as the nonlinearity in the hidden layer and linear activation function is used in the output layer. To find the optimum performance for the training, different number of hidden layer neurons are tested. As seen in Fig.10, the best validation error in the training phase is achieved for 10 neurons in the hidden layer. Hence, NN model for estimating two channel taps for VLC has been obtained.
\begin{figure}[!t]
	\centering
	\includegraphics[width=85mm]{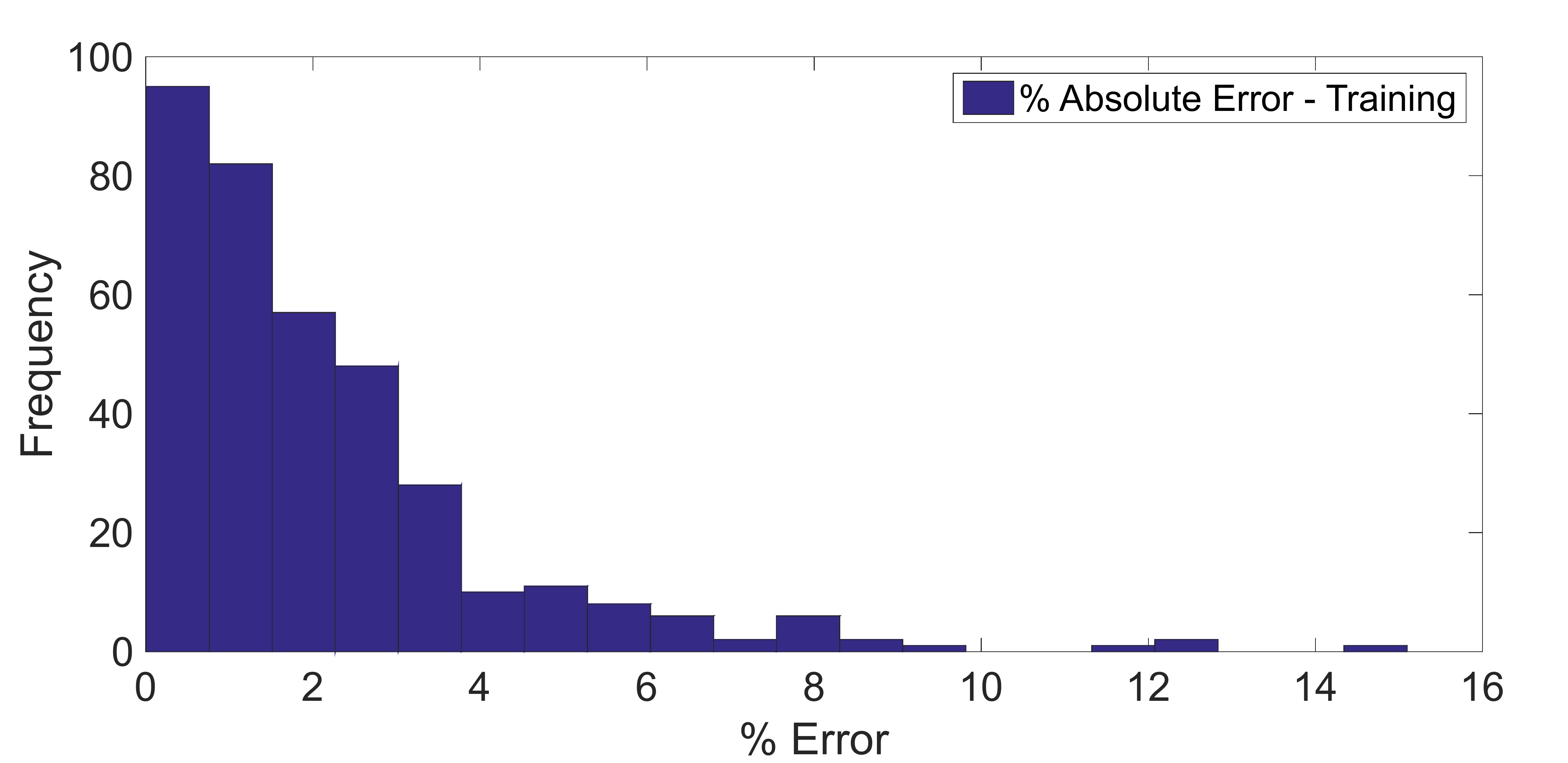}
	\caption{Percentage absolute error distribution for h1 in training data}
	\label{f777}
\end{figure}
\begin{figure}[!t]
	\centering
	\includegraphics[width=85mm]{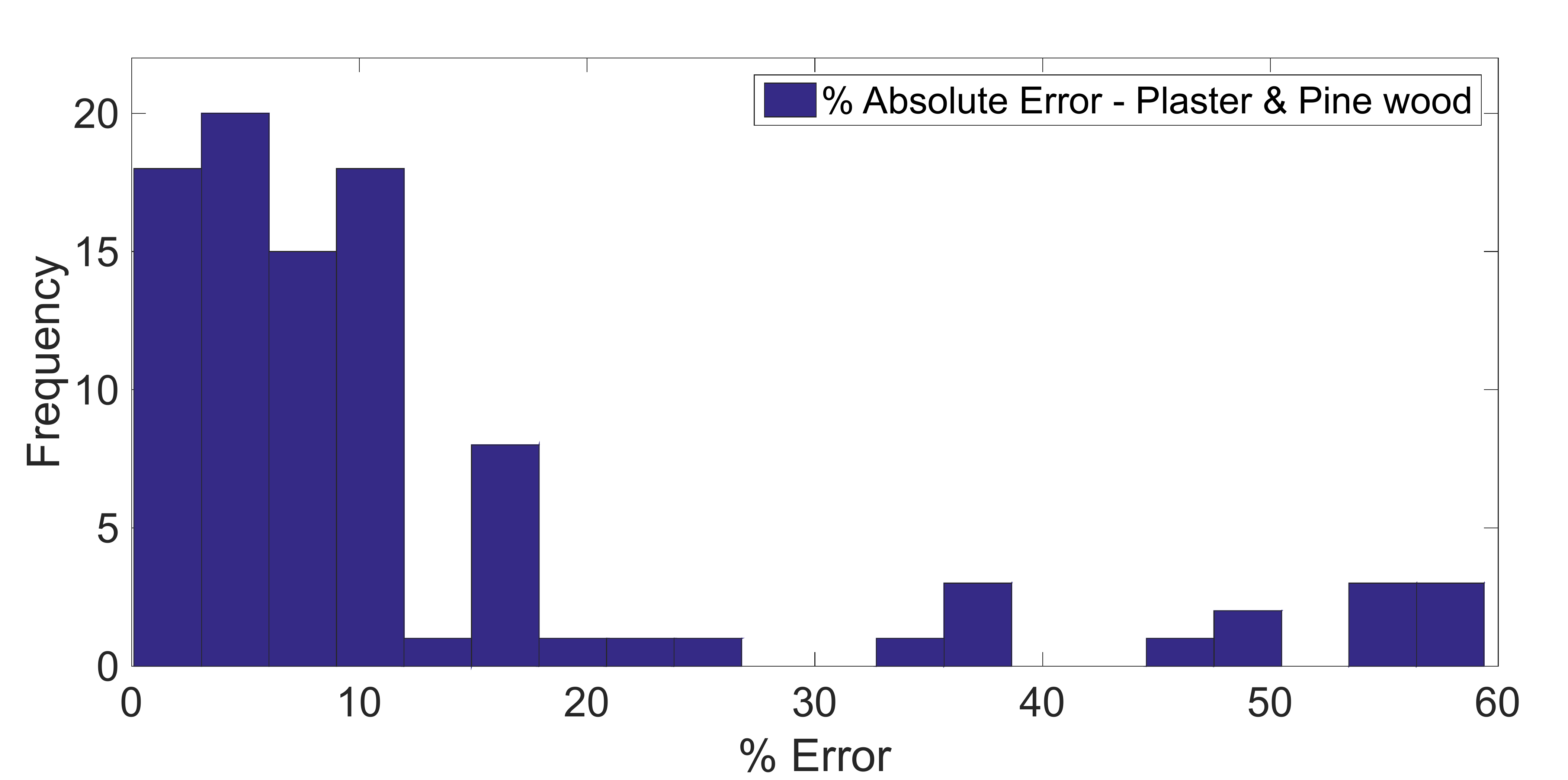}
	\caption{Percentage absolute error distribution for h1 in plaster\&pine-wood data}
	\label{key}
\end{figure}
\begin{table}[!b]
	\centering
	\caption{Mean \% Absolute Error}
	\label{my-label}
	\resizebox{9cm}{!}{
		\renewcommand{\arraystretch}{1.5} 
		\begin{tabular}{|c|c|c|c|}
			\hline
			\textbf{Channel Tap /  Phases} & \textbf{Training} & \textbf{\begin{tabular}[c]{@{}c@{}}Test\\ (Black Flat Paint - Pine wood)\end{tabular}} & \textbf{\begin{tabular}[c]{@{}c@{}}Test\\ (Plaster - Pine wood)\end{tabular}} \\ \hline
			\textbf{h1} & 2.3 & 13.3 & 13.0 \\ \hline
			\textbf{h2} & 1.2 & 4.4 & 2.7 \\ \hline
		\end{tabular}
	}
\end{table}
\begin{table}[!b]
	\centering
	\caption{Average reflectances of hybrid materials}
	\label{my-label}
	\resizebox{7cm}{!}{
		\renewcommand{\arraystretch}{1.4} 
		\begin{tabular}{|c|c|c|}
			\hline
			\textbf{Materials / LED} & \textbf{White} & \textbf{Blue} \\ \hline
			\textbf{Plaster - Pine Wood} & 0.2705 & 0.6274 \\ \hline
			\textbf{Black Flat Paint - Pine Wood} & 0.2445 & 0.5913 \\ \hline
		\end{tabular}
	}
\end{table}
Using two hybrid surfaces and considering other parameters, 8 measurements are carried out randomly at distances between $20cm$ and  $200cm$. Again, each measurement is repeated 10 times. The aim of these measurements is to verify the prediction capability of the trained neural network. Test results are given for the constructed hybrid surfaces in Table III and Table IV. Table III displays MSE performance for the two test cases. As can be seen, the MSE performance is one to two orders of magnitude worse than training. However, as can be seen in Table IV, the mean percent absolute errors for the test cases are still acceptable. Especially, for h2 (transmission with reflections) the mean percentage absolute error is below 5\% for test cases. For a detailed analysis of the percentage absolute error, Figs. 11-12 are given. Here the histograms of percentage absolute errors are given for the training and for one of the test data.
\section{Conclusion}
In this work, we created an experimental setup to estimate VLC channel taps by using neural networks. Based on the knowledge of transmission bandwidth, 2 channel taps are enough to model VLC channel. Six input features (reflectivity of different materials, transmitter types, LOS/NLOS, noise levels, receiver gain and distance between the transmitter and receiver) are used to predict the two channel taps. Experimental data are used to train the MLP network. The results showed that the system can learn the channel taps with 2.3\% mean absolute error in the measurement data set. The channel taps for different hybrid materials in the test phase are predicted with approximately 14\% mean absolute error for tap 1 and 4.3\% for tap 2. Those results indicate that such a procedure may be used effectively to predict channel parameters for VLC. As an alternative to expensive and time consuming simulation softwares, these methods can be used effectively for channel estimation in the VLC. For the future work, different types of surfaces and higher order NLOS channel taps for higher bandwidths will be investigated. Performance of the system in terms of training, validation and test data shows that VLC channel estimation by using neural networks is a promising field for future research.

%
\bibliographystyle{IEEEtran}
\bibliography{references}

\end{document}